\documentclass{article}

\usepackage{arxiv}

\usepackage[utf8]{inputenc} 
\usepackage[T1]{fontenc}    
\usepackage{hyperref}       
\usepackage{url}            
\usepackage{booktabs}       
\usepackage{amsfonts}       
\usepackage{nicefrac}       
\usepackage{microtype}      
\usepackage{lipsum}
\usepackage{graphicx}
\usepackage{color}
\usepackage{caption}
\usepackage{amsmath}
\usepackage{multirow} 
\usepackage{array, makecell}

\newcommand{\rot}[1]{   \rotatebox{75}{#1}}

\title{Using natural language processing techniques to extract information on the properties and functionalities of energetic materials from large text corpora}

\author{
	Daniel C.~Elton\\
	University of Maryland, College Park\\
	Dept.\ of Mechanical Engineering\\
	College Park, MD 20742 \\
	\AND
	Dhruv Turakhia\\
	University of Maryland, College Park\\
	Dept.~of Mechanical Engineering\\
	College Park, MD 20742
	\And
	Nischal Reddy \\
	University of Maryland, College Park\\
	Dept.\ of Mechanical Engineering\\
	College Park, MD 20742
	\AND
	Zois Boukouvalas\\
	University of Maryland, College Park\\
	Dept.~of Mechanical Engineering\\
	College Park, MD 20742
	\And
	Mark D.~Fuge\\
	University of Maryland, College Park\\
	Dept. of Mechanical Engineering\\
	College Park, MD 20742 
	\AND
	~~~~~Ruth M.~Doherty\\
	~~~~~Energetics Technology Center\\
	~~~~~Indian Head, MD 20640
	\And
	~~~~~~~~Peter W.~Chung\\
	~~~~~~~~University of Maryland, College Park\\
	~~~~~~~~Dept.~of Mechanical Engineering\\
	~~~~~~~~College Park, MD 20742 \\
	~~~~~~~~\texttt{pchung15@umd.edu} 
}

\begin{document}
\maketitle

\begin{abstract}
The number of scientific journal articles and reports being published about energetic materials every year is growing exponentially, and therefore extracting relevant information and actionable insights from the latest research is becoming a considerable challenge. In this work we explore how techniques from natural language processing and machine learning can be used to automatically extract chemical insights from large collections of documents. We first describe how to download and process documents from a variety of sources - journal articles, conference proceedings (including NTREM), the US Patent \& Trademark Office, and the Defense Technical Information Center archive on archive.org. We present a custom NLP pipeline which uses open source NLP tools to identify the names of chemical compounds and relates them to function words (``underwater'', ``rocket'', ``pyrotechnic'') and property words (``elastomer'', ``non-toxic''). After explaining how word embeddings work we compare the utility of two popular word embeddings - word2vec and GloVe. Chemical-chemical and chemical-application relationships are obtained by doing computations with word vectors. We show that word embeddings capture latent information about energetic materials, so that related materials appear close together in the word embedding space. 
\end{abstract}

\section{Introduction}
The number of scientific journal articles and reports being published every year is growing exponentially -- since 1945, global scientific output has doubled roughly every nine years \cite{RN300}. This growth has been accompanied by a growing demand for automatable technologies that can be used to keep abreast of developments in science and technology (S\&T). The need for such capabilities is particularly acute for businesses and defense agencies looking to maintain a competitive advantage. Data mining, natural language processing (NLP), and natural language understanding (NLU) are three overlapping areas of research where techniques are being developed to automatically process large volumes of text and extract useful insights. 

Developments in NLP have a rich history extending over many decades, particularly in the biological and medical domains \cite{RN304,RN326,RN328,RN298,RN334}. For many niche NLP tasks shared benchmark datasets are now available containing extensive hand annotations. Creating such datasets is highly resource intensive, however. For example, the creation of the training data for the {\it BioCreative V} Chemical Disease Relation task \cite{RN318} required hand-annotating chemicals, diseases, and chemical-disease interactions in 1,500 PubMed articles. This required a team of subject matter experts using specialized annotation software \cite{RN343}. Similar large scale hand labeling efforts have been undertaken for chemical named entity recognition (NER). For instance, the CHEMDNER corpus is a collection of 84,355 chemical entity mentions from 10,000 PubMed articles labeled and classified manually by a team of experts \cite{krallinger2015chemdner}. Court et al. developed a system which automatically extracts data on the Curie and N{\'e}el temperatures of materials from journal articles \cite{RN306}.

Thus, hand-labeled NLP datasets are important resources that serve well those communities that possess them. However, algorithms trained on hand labeled datasets often lack transferability to other datasets and problems. In this work we explore what can be achieved using word embeddings and off-the-shelf NER tools without having to invest large resources in hand labeling. This is a first effort to assess the viability of NLP methods to assisting in energetics research.

In the context of energetics, we particularly seek an automated system which can extract chemical-chemical, chemical-property, and chemical-application  associations for chemicals used in energetic devices (including binders, energetics, and fuel). We note that relatively little work has been done on extracting chemical-property or chemical-application relations from text. Kim et al. have built a system to extract materials synthesis procedures. A technique they developed that is relevant to our work is what they call ``contextual neighbor sampling'', which uses the word2vec word embedding method (discussed below) \cite{RN313,RN314}. Kim et al. were interested in training an autoencoder network to compress synthesis procedures to a low dimensional space. They were primarily interested in synthesis procedures for a narrow set of materials-BaTiO$_3$ and SrTiO$_3$. However, only about 200 synthesis descriptions were available for those materials, which was not enough to train an autoencoder. To augment their dataset, they used {\it word2vec} embeddings to identify inorganic oxides that appeared in similar contexts. By using the materials that were closest to BaTiO$_3$ and SrTiO$_3$ in the word embedding space (such as CaTiO$_3$, PbTiO$_3$, etc.) they were able to augment their dataset to contain 1,200 synthesis descriptions. 

In this work, we borrow ideas from the NLP system developed by Cheong et al.\cite{RN303,RN302} which extracts information about the functionalities of mechanical parts by identifying ``{\it artifact - function - energy flow}'' triplets. After first describing word embedding techniques and our NLP pipeline, we demonstrate a system which can extract chemical-chemical, chemical-property, and  chemical-application relations from text related to energetic materials. Due to our use of unlabeled data, we believe our approach has a high potential for transferability. 

\section{Word embedding methods}
Word embedding methods create vectors for each word such that words that often appear nearby in the text appear nearby in the embedding space. Word embeddings are used for many different NLP tasks, which can be broken into two categories - {\it intrinsic} and {\it extrinsic} \cite{RN299,RN294}. Intrinsic tasks extract latent semantic relations that are related to underlying meaning. Examples include analogy completion, synonym detection, and the disambiguation of acronyms \cite{RN320}. By contrast, extrinsic tasks are tasks that are more related to syntax, such as named entity recognition, part-of-speech tagging, and semantic role labeling. We tested two of the most popular word embedding techniques---word2vec and GloVe.

\subsection{word2vec}
The term word2vec is actually an umbrella term for a closely related family of word embedding  algorithms \cite{RN323,RN322}. The original {\it word2vec} authors have published two different objective functions for training {\it word2vec} vectors: skip grams \cite{RN323} and continuous bag of words \cite{RN322}. Each of these objective functions may be employed with a variety of different models. The skip gram objective takes a center word $w_c$ as input and tries to reproduce the $(\log)$ probability of context words:
\begin{equation}
\log p\left(w_{c-m},\dots,w_{c-1},w_{c+1},\dots,w_{c+m}|w_c\right) = \log\left( \prod_{-m\leq j\leq m,j\neq 0} p(w_{c+j}|w_c) \right)
\end{equation}
where $m$ is the (half) size of the contextual window. In equation 1 an assumption of conditional independence was made - it is assumed for simplicity that the presence of each context word is independent of the others.	The output of the model is a vector ${\hat{\bf y}}$ in which each element represents the probability that of that word appearing in the context of the center word. The continuous bag of words model performs the opposite task of skip gram - it tries to predict the probability of a center word given a list of context words. One type of model used for the skip gram objective is what may be called a ``one layer $+$ softmax'' model. This means the model consists solely of a one layer neural network followed by a softmax ``squeezing'' function which produces a probability distribution. When using the skip gram objective, the model works as follows: first, the center word is encoded in a one hot vector ${\bf x}_c$. One hot vectors have one element which is equal to 1 (``hot'') corresponding to a particular word, and all other elements equal to zero. The one hot vector ${\bf x}_c$ is multiplied by a matrix ${\bf W}$ of size $V\times N$, where $V$ is the vocabulary size and $N$ is the word vector size. This generates the word vector ${\bf w}_c$. Then the word vector is multiplied by a different matrix ${\bf W}'$ to obtain a weighting vector ${\bf u}$, which is then turned into a probability vector ${\hat{\bf y}}$ using the softmax function. In summary: 
\begin{align}
	{\bf w}_c & = {\bf W} {\bf x}_c\nonumber \\
	{\bf u} & = {\bf W}'{\bf w}_c \\
	{\hat{\bf y}} & = {\rm softmax}({\bf u}). \nonumber
\end{align}
The softmax function ${\hat{\bf y}} = {\rm softmax}({\bf u})$ is defined element-wise as: 
\begin{equation}
{\hat{\bf y}}_i = \frac{\exp(u_i)}{\sum_{j=1}^{V}\exp(u_j)}.
\end{equation}
A simpler method of generating the probability vector is what may be called the  ``single softmax" model, which operates directly on dot products of word vectors: 
\begin{equation}
{\hat{\bf y}}_i = \frac{\exp(({\bf w}^i)^{\top}{\bf w}^c)}{\sum_{j=1}^{V}\exp(({\bf w}^c)^{\top}{\bf w}^j))}.
\end{equation}
Typically, the distance between the probabilities generated by the model ${\hat{\bf y}}$ and the real world probabilities ${\bf y}$ are computed using the cross-entropy metric:
\begin{equation}
H({\hat{\bf y}},{\bf y}) = -\sum_{j=1}^{V} {\hat{\bf y}}_j\log({\hat{\bf y}}_j).
\end{equation}
The model is trained by minimizing cross entropy, usually using stochastic gradient descent. While the ``one layer $+$ softmax'' and ``single softmax'' models are often introduced for pedagogical purposes, they are rarely used in practice because computation of the softmax function with a large vocabulary size $V$ becomes expensive computationally due to the sum in the denominator. Instead, the {\it word2vec} authors introduced two other techniques: hierarchical softmax and negative sampling, both of which are more efficient. An additional implementation detail which is specific to word2vec is the use of subsampling \cite{RN317}. During subsampling, words that are more frequent than a threshold $t$ are removed with a probability $p$. Subsampling can be thought of as a probabilistic variant of stop word removal.

\subsection{GloVe}
The GloVe (``Global Vector'') word embedding, introduced by Pennington et al. in 2014, takes a very different approach to generating word embeddings \cite{RN327}. In GloVe, word vectors are used to reproduce the co-occurrence matrix $X_{ij}$ as faithfully as possible. Each element in the co-occurrence matrix tabulates the probability (inverse frequency) with which word $i$ appears in the context of word $j$, where context is defined using a window with half width $m$. The GloVe authors use the following model for the co-occurrence matrix:
\begin{equation}
{\bf w}_i^{\top}{\tilde{\bf w}}_j + {\bf b}_i + {\bf b}_j \approx \log(1 + X_{ij}).
\end{equation}
The idea of factorizing the log of the occurrence matrix in such a way is closely related to latent semantic analysis (LSA) \cite{RN351}. The method generates two sets of word vectors ${\bf w}_i$ and ${\tilde{\bf w}}_i$. Typically the left and right contexts are distinguished, so $X_{ij}$ is asymmetric, and these two word vectors are different. A single word vector is obtained as ${\bf w}_i' = {\bf w}_i + {\tilde{\bf w}}_i$. The co-occurrence matrix is sparse, containing only a few values with large probabilities and having most elements being very close to or equal to zero. Therefore, a weighted least squares procedure is used for training the model.

\subsection{Training word embeddings}
The main hyperparameters that need to be tuned when training word embeddings are the embedding dimension $D$ and the (half) window size $m$. Additional parameters that may appear are the frequency cutoff for word consideration in training and the subsampling parameter $t$. Default values in most software packages are $D=100$ or $D=150$ and $m=5$. Setting $D$ too large (ie. $D=1000$) or too small (ie. $D=50$) will significantly degrade performance. Previous hyperparameter studies suggest the optimal dimensionality is typically between $D\approx 200 - 300$ \cite{RN330,RN305}. The optimal setting for $m$ may be more sensitive to the task(s) at hand - one hyperparameter study reports that accuracy on a variety of semantic and syntactic tasks plateaus around $m = 4 - 8$ \cite{RN330} while a different study on biomedical NLP finds a plateau when $m = 16 - 20$ \cite{RN305}. GloVe and word2vec may have different optimal settings. It has been noted that independent hyperparameter optimization usually causes word2vec and GloVe to converge to similar accuracies \cite{RN317}. After testing several different settings and not noticing major qualitative differences we settled on using $D=200$ and $m = 8$. 

\subsection{Measuring word embedding similarity}
The cosine similarity  $S_c\in [-1,1]$ between two word embedding vectors ${\bf w}^1$ and ${\bf w}^2$ is defined as:
\begin{equation}
S_c({\bf w}^1,{\bf w}^2) = \frac{{\bf w}^1\cdot{\bf w}^2}{||{\bf w}^1||||{\bf w}^2||}.
\end{equation}
There are several theoretical motivations for focusing on cosine similarity over Euclidean distance. The first is that it disregards the length of the word vector. We found that the length of our word vectors depends on the number of times the word appears, which has been recognized as a common phenomenon. This is considered an issue since the semantics of words (which word vectors are supposed to encode) should be independent of the number of times that a word appears in a corpus. Cosine similarity is less sensitive to frequency-based effects, although it has been shown that cosine similarities are still contaminated by frequency-based effects to some extent \cite{RN294}. Additionally, it has been shown that when using cosine similarity, fewer ``hubs'' appear in the nearest neighbor lists compared to when Euclidian distance is used \cite{RN329}. Hubs are words that appear in the $k$-nearest neighbor ranking lists for a very large number of words. Hub words are generally words that do not have specific semantic relevance to the query word and thus are not considered useful. The presence of hubs (called ``hubness'') appears to be a near-universal feature of  high dimensional data \cite{RN329}.

\section{Corpus construction and pipeline}
Collecting a corpus of text can be one of the most time consuming parts of any NLP endeavor. Articles in HTML format are preferable to those in PDF because the text of the article’s body is tagged so that headings, captions, tables, and references can easily be ignored during text extraction. However, different online publishers use different HTML conventions, so separate HTML parsers must be developed for each publisher, making this approach labor intensive. Also, most publishers have security mechanisms in place to prevent large-scale web scraping. 

\subsection{Patents}
A benefit of working with patents is that they can be obtained as structured .xml objects, allowing the text to be extracted, as with HTML. In the United States Patent Classification (USPC), the relevant section for energetic materials is class 149, ``Explosive and Thermic Compositions or Charges'', which has a total of 3,136 patents associated with it as either a primary or secondary classification. In the international Cooperative Patent Classification (CPC) the relevant class is C06, ``Explosives, Matches'', in particular subclass B (C06B): ``Explosives or Thermic Compositions; Manufacture Thereof; Use of Single Substances As Explosives'', which is associated with a total of 4,059 US patents. Free bulk downloads of patent data in .xml format are available online \cite{26,27}. A wide variety of open source toolkits for processing these .xml files can be found on GitHub. We used a modified version of the patent-parsing-tools code \cite{28} to automatically download and extract the patent .xml files and then used a custom Python script to search the .xml files and extract text from patents that had C06B as their primary or secondary classification.

\subsection{Proceedings and other sources}
In addition to patents, our dataset consisted of several sets of proceedings papers from the International Detonation Symposium (``IDS'', 1976, 2002, 2014), Fraunhofer Institute for Chemical Technology Annual Conference (``ICT'', 2008, 2010-2018) and NTREM (2014, 2015). Text from NTREM abstracts from 2004-2018 was also included. These pdfs for these proceedings were taken from conference CD-ROMs and the Defense Technical Information Center (DTIC) website, a publicly available repository of unclassified materials run by the US Department of Defense. We used the pdfminer.six Python 3 package to extract text from the pdfs \cite{29}. We also found we could download large sets of documents on energetic materials from the DTIC archive on archive.org using the {\it internetarchive} Python package. However, the text (from optical character recognition) was poor quality and was not used. 

\begin{figure}
	\centering
	\includegraphics[width = 0.7\textwidth]{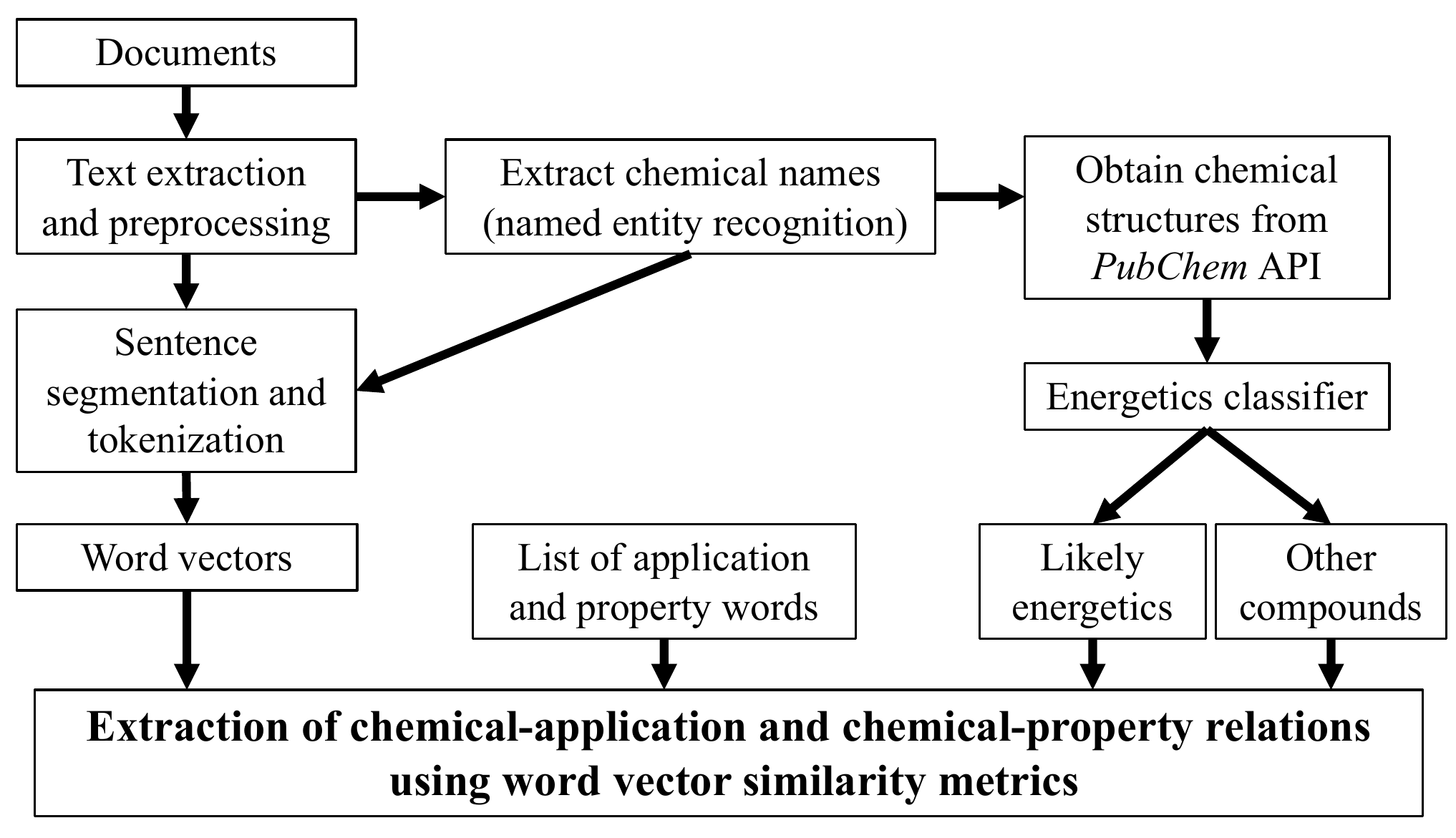}
	\caption{The NLP pipeline used in this work. }
	\label{pipeline}
\end{figure}

\subsection{Proceeding pipeline}
Our NLP pipeline is shown in Figure~\ref{pipeline}. The preprocessing procedure can be summarized as follows:
\begin{itemize}
	\item[1.] Convert .pdfs to text with PDFminer.six or extract text from patent .xml.
	\item[2.] Remove line breaks, references, and if possible metadata, figure captions, and tables.
	\item[3.] Remove stop words using NLTK stop word list combined with a custom stop words list. 
	\item[4.] Remove abbreviations (``Dr.'', ``No.'', etc) that confuse the sentence segmentor using regular expressions. Remove all numbers with decimal places or percentage signs.
	\item[5.] Extract chemical names with {\it ChemDataExtractor}. 
	\item[6.] Call {\it PubChem} application program interface (API) to filter actual chemical names and obtain chemical structures.
	\item[7.] Use a machine learning based classifier to classify energetic chemical names.  
	\item[8.] Remove spaces and dashes from chemical names in text.
	\item[9.] Segment sentences using NLTK {\it sent\_tokenize()}.
	\item[10.] Convert all text to lower case and tokenize sentences using NLTK {\it simple\_preprocess()}.
\end{itemize}

To extract chemical names we used the ChemDataExtractor Python package \cite{RN282}. In a recent comparison, ChemDataExtractor obtained the highest score compared to 10 other models that have been developed for the CHEMDNER ``chemical entity mention'' classification task \cite{RN315}. Another package which achieved a comparable score is the Scientific Natural Language Toolkit (\url{https://github.com/skoblov-lab/scilk}). For each chemical name identified by ChemDataExtractor we made a call to the {\it PubChem} API using the {\it PubChemPy} Python package to verify if it was indeed a chemical name. We also obtained the chemical graphs from PubChem in the form of SMILES strings. Next we used a custom classifier to classify chemicals as energetic vs non-energetic materials. The classifier was trained on a dataset of 617 known energetic materials and 1,128 non-energetic materials which were mostly pesticides and agrochemicals \cite{RN341}. 80\% of the data was used for training and 20\% was used as a hold out test set for evaluation. The chemical structures were featurized using a concatenation of an Estate fingerprint vector with the sum over bonds (bond counting) and custom descriptor set feature vectors. The latter two featurization methods are methods that were developed and tested in previous work where we applied machine learning to predicting the properties of energetic materials \cite{elton2018applying,Barnes2018arxiv}. For classification we used a linear support vector machine (SVM) which achieved an area under the receiver operating characteristic (auROC) score of 0.99, an overall accuracy of 0.98, and a false positive rate of 0.02 on the hold out test set. We also tested a random forest classifier which achieved a similar accuracy. When the SVM classifier was transferred from its original task to classifying chemicals from the corpus, the false positive rate became much higher. Therefore, we tuned the threshold of the classifier by hand to make it more conservative and reduce its false positive rate. 

A weakness of NLP systems based on word embeddings is that chemical names with spaces or dashes get split – for instance ``potassium nitrate'' is considered as two separate words. One solution to this issue, developed by the original {\it word2vec} authors, is a method called phrase skip-gram \cite{RN323}. In our work, however, we choose to circumvent this issue by removing all spaces and dashes from chemical names before sending the text to the tokenizer.

\section{Results}

\subsection{Investigating clustering behavior}
\begin{figure}
	\centering
	\includegraphics[width = 1.02\textwidth]{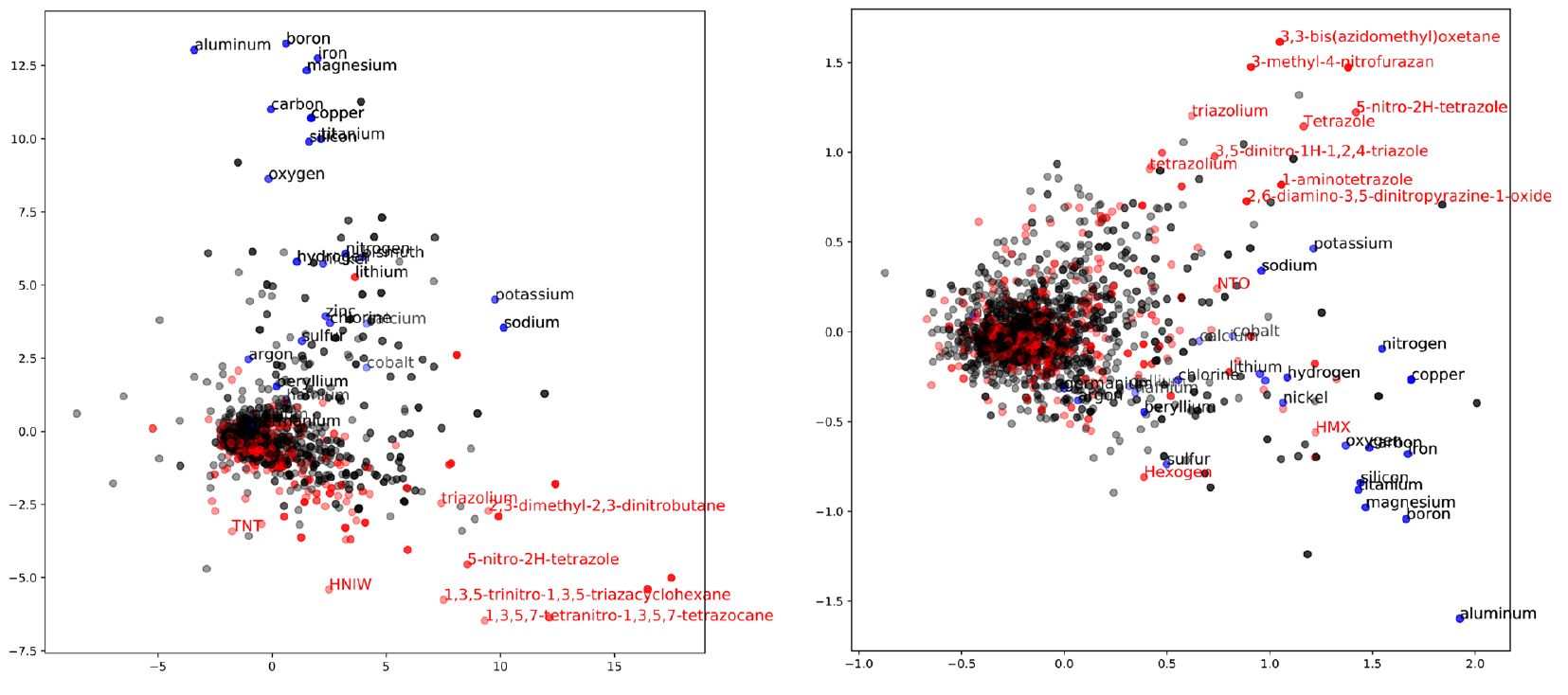}
	\caption{Projection of the 200 dimensional word embeddings along the first two principal components identified by principal components analysis, for word2vec (left) and GloVe (right).}
	\label{projections}
\end{figure}

Table \ref{groupdistances} shows cosine similarities and Euclidean distances between three groups –``energetics'', ``binders'', and ``elements''. The table gives evidence that word embeddings of chemical names are weakly clustered according to which group they belong two. Clustering is indicated by the fact that in all cases intragroup similarities are larger than intergroup similarities. Similar clustering behavior is observed for both word2vec and GloVe. We visualized the clustering of the word vectors for chemical names using several dimensionality reduction techniques – principal components analysis (PCA) \cite{RN350}, t-stochastic neighbor embedding (t-SNE)\cite{RN349}, and spectral embedding. Some small clusters of energetic names were observed in t-SNE (not shown) but they did not appear to be semantically meaningful. Of the three embedding techniques, only PCA preserves the structure of the original space, since PCA amounts to just a rotation of axes. The projection to the first two principal components is shown in Figure \ref{projections}. In both word2vec and GloVe, names corresponding to chemical elements cluster to one side of the space. In both word2vec and GloVe, some names classified as likely energetic materials (shown in red) appear separated from the rest. 

\begin{table*}
	\centering{\begin{tabular}{c|c|c|c|c|c}
		&  \rot{\footnotesize{embedding}}    &  \rot{\footnotesize{groups}}        & \rot{\footnotesize{energetics}}& \rot{\footnotesize{binders}} & \rot{\footnotesize{elements}} \\
		\hline
		\multirow{6}{*}{\rotatebox{90}{\textbf{patents}}}&\multirow{3}{*}{word2vec} &        energetics & \textbf{0.79}, 4.17 &                        &                  \\
		&   & binders  & \textbf{0.50}, 6.50 & \textbf{0.82}, 2.05 &                  \\
		&   & elements & \textbf{0.22}, 10.24 & \textbf{0.38}, 8.25 & \textbf{0.57}, 7.20 \\ \cline{2-6}
		&\multirow{3}{*}{GloVe} &  energetics & \textbf{0.59}, 0.68 &                        &                      \\
		& &  binders   & \textbf{0.32}, 0.98 & \textbf{0.64}, 0.29 &                  \\
		&  &  elements  & \textbf{0.13}, 1.85 & \textbf{0.15}, 1.69 & \textbf{0.64}, 1.30 \\
		\hline
		\multirow{6}{*}{\rotatebox{90}{\textbf{proceedings}}}&\multirow{3}{*}{word2vec}&        energetics & \textbf{0.72}, 6.88 &   &  \\
		& & binders  & \textbf{0.40}, 11.10 & \textbf{0.81}, 3.92 &                  \\
		& & elements & \textbf{0.14}, 13.52 & \textbf{0.27}, 11.10 & \textbf{0.56}, 8.27 \\ \cline{2-6}
		& \multirow{3}{*}{GloVe} &  energetics & \textbf{0.83}, 1.08 &                        &                      \\
		& &  binders   & \textbf{0.59}, 1.80 & \textbf{0.84}, 0.70 &                  \\
		& &  elements  & \textbf{0.27}, 2.26 & \textbf{0.36}, 1.88 & \textbf{0.50}, 1.48 \\
		\hline
	\end{tabular}}
	\caption{ Cosine similarities (bold) and Euclidean distances between groups. The groups were defined as follows: energetics=['hmx', 'tatb', 'rdx', 'petn', 'tnt', 'nto'], binders=['htpb', 'viton',  'wax'], elements=['aluminum', 'carbon', 'zinc', 'nitrogen', 'silicon', 'boron', 'magnesium', 'oxygen', 'hydrogen', 'iron'].}
	\label{groupdistances}
\end{table*}

\subsection{Most common chemical names and chemical-application relations}
Table \ref{chemnames} shows the most common chemical names found in the corpus. We created a list of 80 application related words (such as ``airbag'', ``underwater'', ``nontoxic'', ``nanoparticle'', ``binder'', ``desensitizer''). The most similar application words for each chemical name are shown. Next to each name is shown the total number of times it appeared along with the  number of times that {\it ChemDataExtractor} identified the name as a chemical name. The table reveals that {\it ChemicalDataExtractor} misses identifying a large number of instances for most chemical names. We found that {\it ChemDataExtractor} could only identify chemical names that were embedded in sentences and could not identify chemical names in lists, which partially explains the low detection rate.

\begin{table*}
	\centering{
	\begin{tabular}{c c c c c c} 
		& {\bf chemical name} & ${\bf N}$ & ${\bf N_{{\rm CDE}}}$ & {\bf \textbf{\emph{word2vec}} application words} & {\bf \textbf{\emph{GloVe}} application words} \\ 
		\hline 
		1 & HMX & 5592 & 456 & plastic, explosive & binder, plastic\\ 
		2 & RDX & 5098 & 9 & insensitive, explosive & binder, plastic\\
		3 & AND & 4369 & 5 & blasting, detonator & primary, oxidizer\\
		4 & TNT & 3626 & 304 & explosive, underwater & plastic, explosive\\
		5 & nitrocellulose  & 1960 & 1749 & plasticizer, propellant & binder, plasticizer\\
		6 & TATB & 1890 & 1779 & insensitive, explosive & plastic, insensitive\\
		7 & PETN & 1877 & 1736 & explosive, detonator & plastic, insensitive\\
		8 & 3-nitro-1,2,4-triazol-5-one & 1484 & 13 & primary, secondary & elastomer, thermoplastic\\
		9 & 5-nitro-2H-tetrazole & 1484 & 4 & primary, secondary & elastomer, thermoplastic\\
	\end{tabular}}
	\caption{Most common chemical names in the corpus. $N$ is the total number of times the word appears in the corpus, and $N_{\rm CDE}$ is the number of times the word was identified as a chemical name by {\it ChemDataExtractor}. The two columns on the right give the most similar application words found in the word2vec and GloVe word embedding spaces.} 
	\label{chemnames} 
\end{table*}

\subsection{Discovering chemical-chemical and application-chemical relations}
One of the first tests we ran was to see if our system could detect synonyms. We found that synonyms often appeared in the top-5 similarity rankings for the most similar words. For instance, when using word2vec, for ``RDX'' the third most similar word (after ``HMX'' and ``TNT'') was an older name for RDX, ``hexogen''. Similarly, for PETN, the third most similar word was its full name, ``pentaerythritol tetranitrate''. When only the patent data was used, synonyms often appeared higher in the rankings, which may have been due to the fact that the patent text was cleaner than the proceedings text. Next we investigated if word embedding similarity searches can be used to extract application – chemical relations. Table \ref{rankings} shows some sample results for three application query words - ``rocket'', ``air'', and ``underwater''. Three levels of filtering are applied to the similarity rankings – first no filtering,  then filtering out only chemical names identified by {\it ChemDataExtractor} and found on {\it PubChem}, and finally filtering out only chemical names identified as likely energetic materials by the SVM classifier. The word ``air'' was chosen to find energetic materials which are used for ``aig bags'' in the patent data. Argon and nitrogen, which are both a common gases used as airbag inflators are found high on the list for  word2vec. While many of the words are not related to airbags, semicarbazide nitrate is an experimental energetic material for airbags. By searching through the corpus we determined what various acronyms stand for (this process could be automated). Under the results for the query word ``air'', ``HoB'' stands for ``height of blast'' and ``LAG'' stands for ``liquid assisted grinding'', a processing technique for energetic materials.  In the results for ``underwater'', ``EBX'' stands for ``enhanced blast formulation'', ``TBE'' stands for ``thermobaric explosive'', and ``NOL'' stands for Naval Ordnance Laboratory. In the results for ``rocket'', NOP stands for ``nitrous oxide-propane'', a proposed type of rocket fuel. 

\begin{table*}
	\centering\renewcommand\cellalign{ll}
	{\footnotesize\centering{\begin{tabular}{p{.1cm}|l|l|l|l|l|l|}
			&   \multicolumn{2}{c}{{\bf Similarity ranking}}   &  \multicolumn{2}{c}{{\bf Similarity ranking}}   & \multicolumn{2}{c}{{\bf Similarity ranking}} \\
			
			&   \multicolumn{2}{c}{{\bf for all words}}   &  \multicolumn{2}{c}{{\bf for chemical names}}   & \multicolumn{2}{c}{{\bf for likely energetic materials}} \\
			
			\hline
			
			& ~~~~~~\textbf{\emph{word2vec}} & ~~~~~~~~~~\centering\textbf{\emph{GloVe}} & ~~~~~~~~~~\textbf{\emph{word2vec}} & ~~~~~~~~~~~\textbf{\emph{GloVe}} & ~~~~~~~~\textbf{\emph{word2vec}} & ~~~~~~~~~~\textbf{\emph{GloVe}}\\
			
			\hline
			
			\multirow{9}{*}{\rotatebox{90}{\textbf{underwater}}}& 0.70\hspace{0.1mm} deepwater & 0.79\hspace{0.1mm}	explosions & 0.42\hspace{0.1mm} pentolite & 0.52\hspace{0.1mm}	pentolite & 0.42\hspace{0.1mm}	pentolite & 0.52\hspace{0.1mm}	pentolite \\[1mm]
			& 0.67\hspace{0.1mm} blast & 0.71\hspace{0.1mm}	explosion & 0.42\hspace{0.1mm} TBE & 0.47\hspace{0.1mm}	baratol& 0.39\hspace{0.1mm} Missile& 0.40\hspace{0.1mm} \makecell{Triaminotri-\\nitrobenzene}\\[1mm]
			& 0.64\hspace{0.1mm}	ebx & 0.65\hspace{0.1mm} mitigation & 0.39\hspace{0.1mm} Missile & 0.40\hspace{0.1mm}	\makecell{Triaminotri--\\nitrobenzene}& 0.38\hspace{0.1mm}	NOL &0.35\hspace{0.1mm} \makecell{Hexyl octyl\\decyl adipate}\\[1mm]
			& 0.63\hspace{0.1mm}	mitigation & 0.64\hspace{0.1mm} desensitization & 0.38\hspace{0.1mm} Bullet& 0.35\hspace{0.1mm} \makecell{Hexyl octyl\\decyl adipate} & 0.38\hspace{0.1mm} Trotyl& 0.34\hspace{0.1mm} \makecell{Dinitro-\\glycerine}\\[1mm]
			& 0.62\hspace{0.1mm}	thermobaric & 0.63\hspace{0.1mm}	nfined & 0.38\hspace{0.1mm} NOL & 0.35\hspace{0.1mm} \makecell{Methoxypro-\\pylacetate}& 0.42\hspace{0.1mm} cyclotol& 0.34\hspace{0.1mm}	OCTOL\\[1mm]
			\hline
			
			\multirow{9}{*}{\rotatebox{90}{\textbf{air}}}&0.70\hspace{0.1mm} atmosphere& 0.80\hspace{0.1mm}	bags& 0.63\hspace{0.1mm} Argon& 0.66\hspace{0.1mm}	Argon & 0.26\hspace{0.1mm} HoB & 0.35\hspace{0.1mm} \makecell{Di-n-hexyl\\phthalate} \\[1mm]
			& 0.63\hspace{0.1mm}	Argon & 0.73\hspace{0.1mm} entrapped&0.49\hspace{0.1mm}	Steam & 0.46\hspace{0.1mm} Oxygen & 0.20\hspace{0.1mm} Missile& 0.32\hspace{0.1mm}	\makecell{Hexyl octyl\\decyl adipate}\\[1mm]
			& 0.60\hspace{0.1mm} dust & 0.72\hspace{0.1mm}	bag& 0.48\hspace{0.1mm} \makecell{Carbon\\Dioxide} & 0.45\hspace{0.1mm}	NITROGEN & 0.20\hspace{0.1mm} LAG & 0.30\hspace{0.1mm} \makecell{semicarbazide\\nitrate}\\[1mm]
			& 0.60\hspace{0.1mm} surrounding& 0.70\hspace{0.1mm} entrapping & 0.47\hspace{0.1mm} \makecell{nitrogen\\gas}& 0.41\hspace{0.1mm}	\makecell{Dimethoxy-\\ethane} & 0.20\hspace{0.1mm} pentolite & 0.29\hspace{0.1mm} \makecell{Diethylhexyl \\adipate}\\[1mm]
			& 0.59\hspace{0.1mm}	bag& 0.67\hspace{0.1mm} surrounding& 0.46\hspace{0.1mm}	cyclone& 0.37\hspace{0.1mm} ALUMINIUM & 0.19\hspace{0.1mm} tetradecane& 0.29\hspace{0.1mm}	\makecell{Diisobutyl\\adipate}\\[1mm]
			\hline
			
			\multirow{9}{*}{\rotatebox{90}{\textbf{rocket}}}& 0.84\hspace{0.1mm} motors & 0.9\hspace{0.1mm} motors & 0.65\hspace{0.1mm} rocket fuel& 0.63\hspace{0.1mm} rocket fuel& 0.65\hspace{0.1mm}	rocket fuel & 0.63\hspace{0.1mm}	rocket fuel\\[1mm]
			& 0.81\hspace{0.1mm}	motor & 0.89\hspace{0.1mm} engines & 0.65\hspace{0.1mm}	Propel & 0.54\hspace{0.1mm} Missile& 0.51\hspace{0.1mm} Missile & 0.54\hspace{0.1mm}	Missile\\[1mm]
			& 0.81\hspace{0.1mm} engines & 0.87\hspace{0.1mm} engine & 0.51\hspace{0.1mm} Missile & 0.54\hspace{0.1mm}	NEM & 0.46\hspace{0.1mm} RAP & 0.31\hspace{0.1mm}	\makecell{Di-n-hexyl\\phthalate}\\[1mm]
			& 0.78\hspace{0.1mm}	ducted & 0.86\hspace{0.1mm} biblarz & 0.47\hspace{0.1mm} \makecell{AMMONIUM\\PERCHLORATE }& 0.42\hspace{0.1mm} blue line& 0.31\hspace{0.1mm} \makecell{NITRO-\\GLYCERINE} & 0.29\hspace{0.1mm} \makecell{NITRO-\\GLYCERINE}\\[1mm]
			& 0.75\hspace{0.1mm} hybrid & 0.86\hspace{0.1mm} nop & 0.46\hspace{0.1mm}	RAP& 0.41\hspace{0.1mm} SRP& 0.29\hspace{0.1mm}	\makecell{nitrogen\\tetroxide} & 0.29\hspace{0.1mm} TEGDN\\[1mm]
			\hline
	\end{tabular}}}
	\caption{Similarity rankings for three example query words – ``underwater'', ``air'', and ``rocket''.}
	\label{rankings}
\end{table*}

\section{Conclusion}
In this paper we have presented a proof of concept system demonstrating how chemical-chemical and application-chemical relations can be obtained from large text corpora, without the need for hand labeling. We first compiled a corpus of text comprised of open energetics  documents. Using natural language processing techniques based on word2vec and GloVe word embeddings we demonstrated how our system can extract chemically meaningful associations and provide novel insights into how chemicals are used in energetic materials research and energetic systems.

\subsubsection*{Acknowledgments}
Support for this work is gratefully acknowledged from the U.S.~Office of Naval Research under grant number N00014-17-1-2108 and from the Energetics Technology Center under project number 2044-001. Partial support is also acknowledged from the Center for Engineering Concepts Development in the Department of Mechanical Engineering at the University of Maryland, College Park. We thank Dr.~Elan Moritz and Dr.~Bill Wilson from the Energetics Technology Center for their encouragement and advice. 

\bibliographystyle{unsrt}
\bibliography{references}

\end{document}